\ifdefined\pdfoutput\pdfoutput=1\fi
\RequirePackage[T1]{fontenc}
\documentclass[letterpaper,10pt,conference]{ieeeconf}
\ifdefined\pdfminorversion\pdfminorversion=7\fi

\IEEEoverridecommandlockouts   %
\usepackage{amsmath,amssymb,amsfonts}
\usepackage{graphicx}
\usepackage{booktabs}
\usepackage{multirow}
\usepackage{xcolor}
\usepackage{needspace}
\usepackage{flafter}                    %
\usepackage{placeins}                   %
\usepackage{cite}                        %
\usepackage[capitalize]{cleveref}        %

\crefname{figure}{Fig.}{Figs.}
\Crefname{figure}{Fig.}{Figs.}
\crefname{table}{Table}{Tables}
\Crefname{table}{Table}{Tables}
\crefname{section}{Section}{Sections}
\Crefname{section}{Section}{Sections}
\creflabelformat{equation}{(#2#1#3)}
\crefname{equation}{}{}
\Crefname{equation}{}{}
\crefname{appendix}{Appendix}{Appendices}
\Crefname{appendix}{Appendix}{Appendices}

\graphicspath{{figures/}}

\newcommand{\wmppo}{WM-LOCO}
\newcommand{\ppo}{PPO}

\makeatletter
\newcommand{\figurecaptype}{\def\@captype{figure}}
\makeatother

\newcommand{\wmblock}[1]{\par\noindent\textbf{#1.}\enspace\ignorespaces}
\newcommand{\wmmodule}[2]{\par\noindent #1) #2:\enspace\ignorespaces}
\renewcommand{\thesubsectiondis}{\kern0.5bp\Alph{subsection}.}

\newcommand{\diffEasy}{Easy}
\newcommand{\diffMed}{Medium}
\newcommand{\diffHard}{Hard}

\newcommand{\gapWidthEasy}{\ensuremath{0.40\text{--}0.55}}
\newcommand{\gapWidthMed}{\ensuremath{0.90\text{--}1.10}}
\newcommand{\gapWidthHard}{\ensuremath{1.80\text{--}2.00}}

\newcommand{\stairRiserE}{\ensuremath{9.0\text{--}10.0}}
\newcommand{\stairRiserM}{\ensuremath{12.5\text{--}13.5}}
\newcommand{\stairRiserH}{\ensuremath{18.5\text{--}19.5}}
\newcommand{\stairTreadE}{25.0}
\newcommand{\stairTreadM}{25.0}
\newcommand{\stairTreadH}{25.0}
\newcommand{\stairStepsE}{3}
\newcommand{\stairStepsM}{4}
\newcommand{\stairStepsH}{5}
\newcommand{\stoneEdgeE}{\ensuremath{28.7\text{--}29.3}}
\newcommand{\stoneEdgeM}{\ensuremath{26.7\text{--}27.3}}
\newcommand{\stoneEdgeH}{\ensuremath{23.3\text{--}24.0}}
\newcommand{\stoneGapE}{\ensuremath{40.7\text{--}41.3}}
\newcommand{\stoneGapM}{\ensuremath{42.7\text{--}43.3}}
\newcommand{\stoneGapH}{\ensuremath{46.0\text{--}46.7}}

\newcommand{\realBeamLength}{\ensuremath{3\,\mathrm{m}}}
\newcommand{\realBeamWidth}{\ensuremath{0.10\,\mathrm{m}}}

\newcommand{\realStairRiserCm}{15}
\newcommand{\realStairTreadCm}{25}
\newcommand{\realStairStepsPerFlight}{4}
\newcommand{\realStoneEdgeCm}{25.0}
\newcommand{\realStoneGapCm}{45.0}

\newcommand{\stairStrideEasyPPO}{0.45}  \newcommand{\stairStrideEasyWM}{0.52}
\newcommand{\stairStrideMedPPO}{0.42}   \newcommand{\stairStrideMedWM}{0.53}
\newcommand{\stairStrideHardPPO}{0.40}  \newcommand{\stairStrideHardWM}{0.54}
\newcommand{\stairSpmEasyPPO}{4.91}     \newcommand{\stairSpmEasyWM}{4.45}
\newcommand{\stairSpmMedPPO}{5.04}      \newcommand{\stairSpmMedWM}{4.26}
\newcommand{\stairSpmHardPPO}{5.15}     \newcommand{\stairSpmHardWM}{4.07}
\newcommand{\stairCotEasyPPO}{1.01}     \newcommand{\stairCotEasyWM}{0.95}
\newcommand{\stairCotMedPPO}{1.21}      \newcommand{\stairCotMedWM}{1.01}
\newcommand{\stairCotHardPPO}{1.43}     \newcommand{\stairCotHardWM}{1.14}
\newcommand{\stairPelaccEasyPPO}{6.96}  \newcommand{\stairPelaccEasyWM}{5.28}
\newcommand{\stairPelaccMedPPO}{8.26}   \newcommand{\stairPelaccMedWM}{5.85}
\newcommand{\stairPelaccHardPPO}{9.89}  \newcommand{\stairPelaccHardWM}{6.58}
\newcommand{\stairAraEasyPPO}{3.94}     \newcommand{\stairAraEasyWM}{2.61}
\newcommand{\stairAraMedPPO}{6.19}      \newcommand{\stairAraMedWM}{2.88}
\newcommand{\stairAraHardPPO}{8.14}     \newcommand{\stairAraHardWM}{3.32}
\newcommand{\stairTqEasyPPO}{0.56}      \newcommand{\stairTqEasyWM}{0.59}
\newcommand{\stairTqMedPPO}{0.62}       \newcommand{\stairTqMedWM}{0.63}
\newcommand{\stairTqHardPPO}{0.67}      \newcommand{\stairTqHardWM}{0.66}

\newcommand{\klDyn}{0.5}              %
\newcommand{\klRep}{0.1}              %
\newcommand{\histLen}{5}
\newcommand{\numEnvs}{8192}

\newcommand{\trainGpu}{a single NVIDIA RTX~5880-Ada (48~GB)}

\title{\LARGE \bf World-Model-Augmented Visual Locomotion for Humanoids on Foothold-Constrained Terrain}

\author{Yuxi Liu$^{1,*}$, Lijun Han$^{1,*}$, Ziming Wang$^{1,2}$, Ao Zhang$^{1,3}$, Cong Yang$^{4}$, and Wei Sui$^{1,\dagger}$%
\thanks{$^{*}$Equal contribution. $^{\dagger}$Project lead and corresponding author: {\tt\small wei.sui@d-robotics.cc}}%
\thanks{$^{1}$D-Robotics. $^{2}$Beijing University of Posts and Telecommunications. $^{3}$Harbin Institute of Technology. $^{4}$Soochow University.}%
}

\newsavebox{\teaserbox}
\begin{document}
\raggedbottom

\makeatletter
\begin{lrbox}{\teaserbox}%
\begin{minipage}{\textwidth}
\centering
\noindent
\begin{minipage}[t]{\dimexpr\linewidth/3\relax}
  \centering
  \includegraphics[width=\linewidth,height=0.8\linewidth]{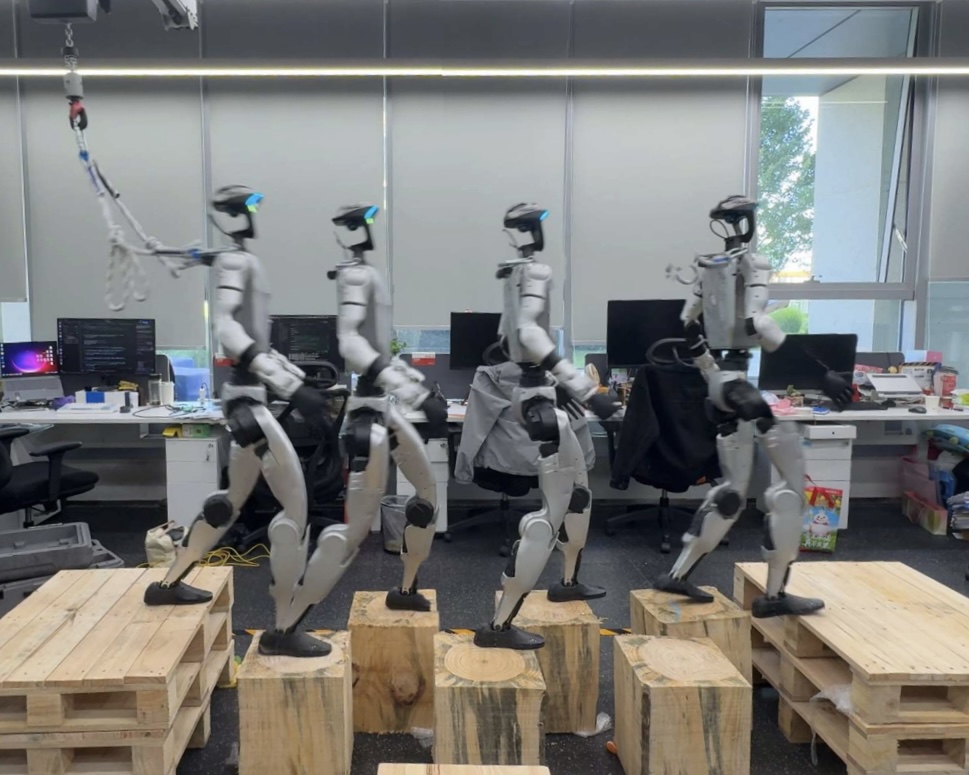}
  \par\vspace{4pt}\small (a) Stepping stones
\end{minipage}%
\begin{minipage}[t]{\dimexpr\linewidth/3\relax}
  \centering
  \includegraphics[width=\linewidth,height=0.8\linewidth]{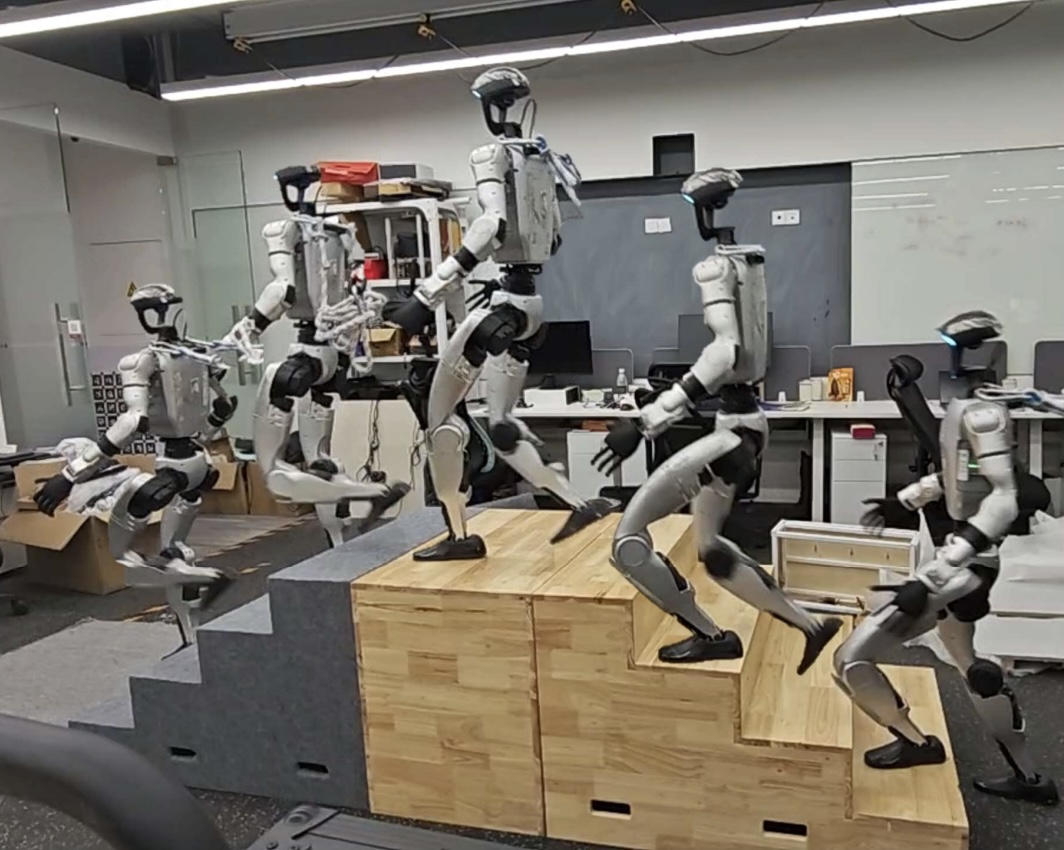}
  \par\vspace{4pt}\small (b) Stairs
\end{minipage}%
\begin{minipage}[t]{\dimexpr\linewidth/3\relax}
  \centering
  \includegraphics[width=\linewidth,height=0.8\linewidth]{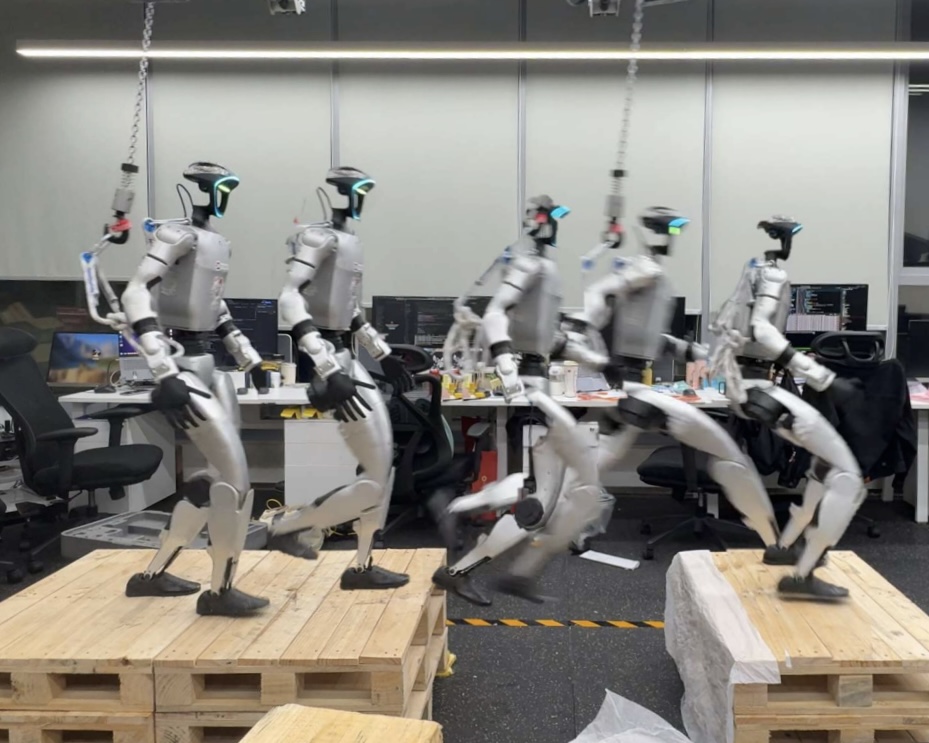}
  \par\vspace{4pt}\small (c) Gap
\end{minipage}\par
\vspace{4pt}
\includegraphics[width=\linewidth]{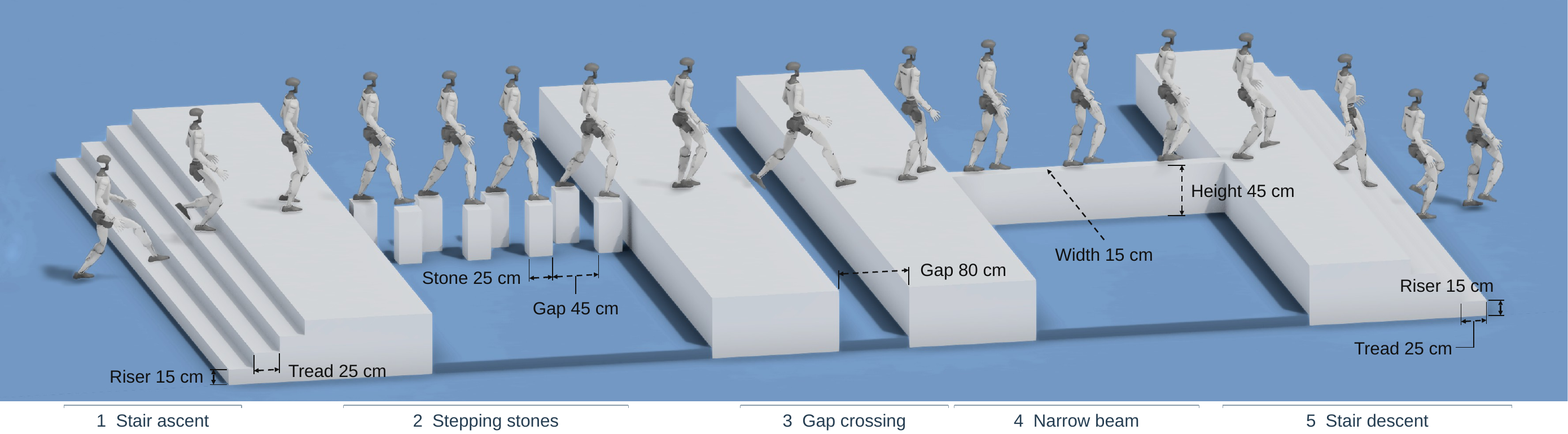}
\par\vspace{2pt}\small (d) Combined course\par
\def\@captype{figure}
\caption{WM-LOCO traverses (a) stepping stones, (b) stairs, and (c) a gap on a Unitree G1 humanoid, and (d) a combined course of stairs, stepping stones, a gap, and a balance beam in simulation. The policy uses proprioception and a single onboard depth stream, without offboard perception or a terrain map.}
\label{fig:realworld_teaser}
\end{minipage}%
\end{lrbox}
\let\originalmaketitle\@maketitle
\def\@maketitle{\originalmaketitle\noindent\usebox{\teaserbox}\par}
\makeatother
\maketitle
\thispagestyle{empty}
\pagestyle{empty}

\begin{abstract}
Foothold-constrained terrain limits feasible footholds to sparse, discontinuous, or narrow support regions, as in gaps, stepping stones, stairs, and balance beams. In such settings, a single misplaced step can cause a fall. Reliable traversal therefore requires precise foot placement and anticipation of upcoming terrain. To address this challenge, we present World-Model-Augmented Visual Locomotion (WM-LOCO), which augments a visual locomotion policy with predictive context from a jointly trained world model. Given onboard depth observations and proprioception, a recurrent state-space model (RSSM) learns action-conditioned latent dynamics that predict future proprioceptive states, egocentric observations, and task rewards. The resulting recurrent feature is then fed into the proximal policy optimization (PPO) actor--critic, providing predictive context for anticipatory foot placement on foothold-constrained terrain. WM-LOCO achieves average success rates of 96.0\% on gaps, 96.9\% on stepping stones, 94.0\% on stairs, and 92.1\% on balance beams, outperforming the evaluated baselines while improving gait efficiency and smoothness. Furthermore, WM-LOCO achieves zero-shot sim-to-real transfer to a Unitree G1, with 92.5\% success across these four terrain classes.
\end{abstract}

\section{Introduction}
\label{sec:intro}
\begin{figure*}[!t]
\centering
\includegraphics[width=\textwidth]{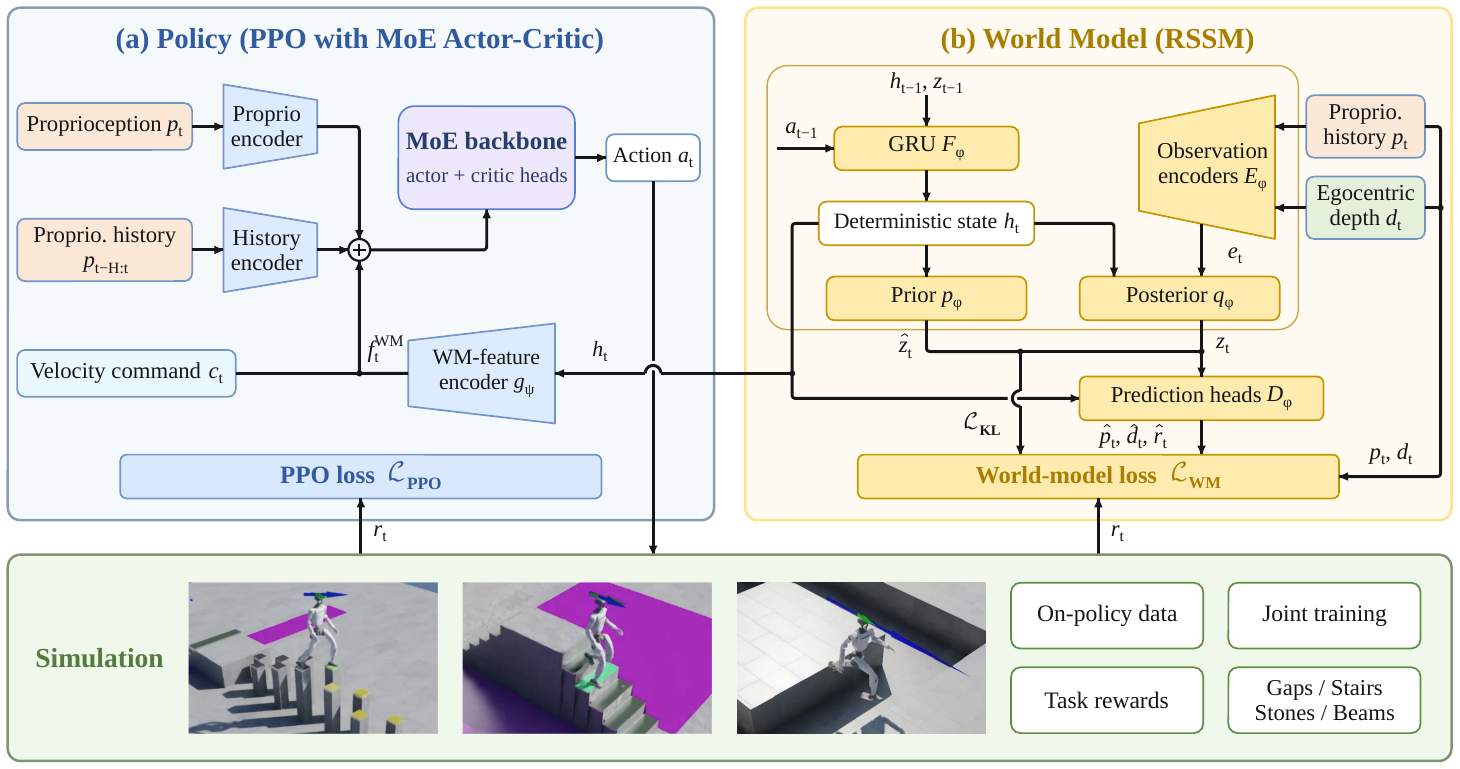}
\caption{WM-LOCO jointly trains an RSSM and a PPO actor--critic. The recurrent feature supplies predictive context to the policy.}
\label{fig:arch}
\end{figure*}

Reinforcement learning has enabled robust locomotion on continuous terrain, where dense support regions allow small foot-placement errors to be corrected within the same step or in subsequent steps~\cite{miki2022learning,radosavovic2024humanoid}. This tolerance diminishes when stable contact is restricted to sparse, discontinuous, or geometrically constrained regions~\cite{start2025}. On such terrain, a foot landing outside the available support can directly cause a fall. We refer to such environments as foothold-constrained terrain. Common examples include gaps, stepping stones, narrow stair treads, and balance beams~\cite{agarwal2023egocentric,duan2022steppingstones}. Successful locomotion on such terrain therefore requires not only precise foot placement, but also anticipation of upcoming support conditions.

Recent humanoid systems~\cite{zhuang2024humanoidparkour,sun2026dpl} often rely on complex or auxiliary mechanisms, such as multi-stage training, teacher--student distillation, explicit foothold supervision, and online planning. BeamDojo adopts a multi-stage training scheme with separate critics for dense locomotion rewards and sparse foothold rewards~\cite{beamdojo2025}. RPL trains privileged expert policies and distills them into a depth-based student policy~\cite{rpl2026}. PLANC combines reduced-order references with multi-stage teacher--student distillation~\cite{planc2026}. Model-based approaches instead perform online footstep planning with mixed-integer quadratic programming~\cite{disconnectedfootholds2026}. This leaves open whether a method exists for learning foothold-constrained locomotion through single-stage, end-to-end training without relying on such complex mechanisms.

\begin{figure*}[!t]
\centering
\includegraphics[width=\textwidth]{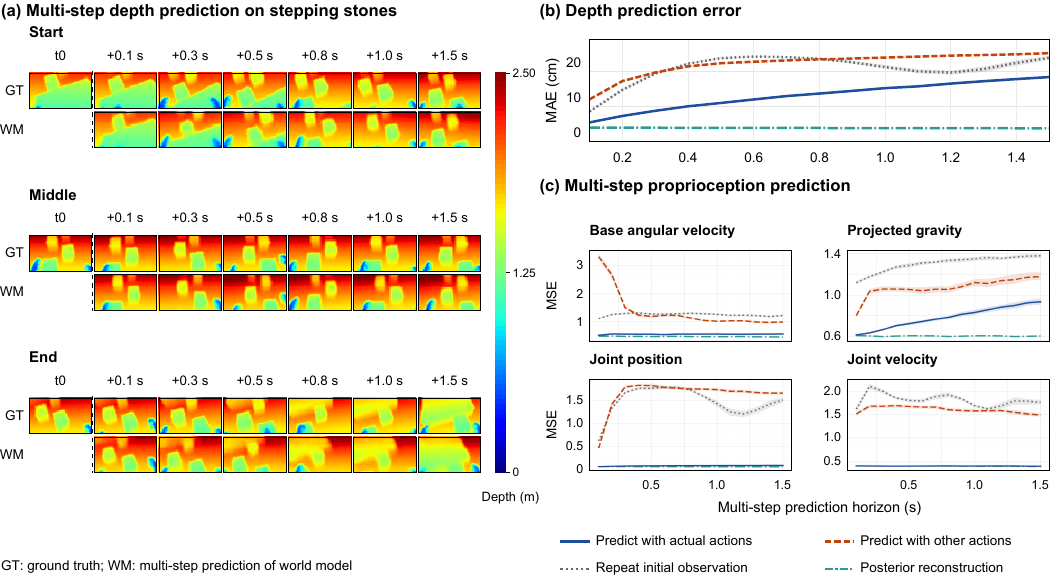}
\caption{Multi-step world-model predictions on stepping stones: \textbf{(a)} depth sequences; \textbf{(b,c)} depth and proprioceptive errors. Rollouts receive no future observations.}
\label{fig:wm-openloop}
\end{figure*}

World models offer a mechanism for learning predictive temporal representations from interaction~\cite{hafner2023dreamerv3,hafner2021dreamerv2}. Prior work has explored state-representation learning for humanoid locomotion, including world-state reconstruction in WMR~\cite{wmr2025}, denoising world-model learning in DWL~\cite{gu2024dwl}, and state prediction in PIM~\cite{pim2024}. Predictive visual representations have also been explored in legged locomotion~\cite{wmp2024}, while other predictive models synthesize future motion rather than model onboard visual dynamics~\cite{dreampolicy2025}. However, few prior humanoid locomotion studies have developed world models that can perform multi-step rollouts given policy actions~\cite{li2025rwm}. Most existing models also exclude visual observations from their predictive scope~\cite{gu2024dwl,wmr2025,li2025rwm}, and thus do not provide future visual context that is particularly important for foothold-constrained terrain.

We therefore introduce World-Model-Augmented Visual Locomotion (WM-LOCO), a single-stage framework that couples proximal policy optimization (PPO)~\cite{schulman2017proximal} with a recurrent state-space model (RSSM) as the world model. The world model encodes onboard depth and proprioception into a recurrent latent state and learns action-conditioned dynamics through observation reconstruction, reward prediction, and prior--posterior matching. These learned dynamics model the evolution of robot state, egocentric observations, and task outcomes, and can be recursively propagated over multiple future steps. The resulting recurrent memory therefore captures predictive structure about how the robot and its egocentric view evolve under its actions. WM-LOCO feeds this memory directly to the PPO actor--critic, giving the policy access to the learned relationship between current motion and upcoming terrain. This predictive context allows the actor to coordinate body motion with upcoming support conditions and prepare foot placement before contact is made.

Simulation and hardware experiments evaluate WM-LOCO across four terrain classes with distinct foothold constraints: stairs, gaps, stepping stones, and balance beams. Multi-step rollout experiments demonstrate that the learned world model can recursively predict the action-conditioned evolution of egocentric observations and robot states without access to future observations. In simulation, WM-LOCO achieves a mean success rate of 94.8\%, averaged equally across the four terrain classes and three terrain difficulty levels---Easy, Medium, and Hard. \Cref{app:terrains} specifies the terrain dimensions used in the evaluation. WM-LOCO achieves the highest mean success rate among the evaluated methods, while also improving gait efficiency and smoothness. On a physical Unitree G1 humanoid, WM-LOCO achieves zero-shot sim-to-real transfer with 92.5\% success across all four terrain classes, using only onboard proprioception and a single depth stream.

\section{Method}
\label{sec:method}
WM-LOCO jointly trains a recurrent world model and a PPO actor--critic (\Cref{fig:arch}).
The world model maintains a recurrent latent state by integrating onboard depth and proprioception, and learns action-conditioned dynamics that predict future observations and rewards over multiple steps. The world model's recurrent memory provides the actor--critic with predictive context, supporting anticipatory foot placement on foothold-constrained terrain.

\subsection{Recurrent World Model}
Our RSSM~\cite{hafner2019planet} combines observation encoders, recurrent latent dynamics, and prediction heads to infer the robot's state from onboard depth and proprioception and predict its evolution under a supplied action sequence. It is trained jointly with PPO through observation reconstruction, reward prediction, and prior--posterior matching.

\wmblock{State inference}
Let $t$ denote the time index and $\phi$ the world-model parameters. Consecutive indices in the world-model recurrence refer to successive model updates. An observation-conditioned state comprises deterministic recurrent memory $h_t$ and a discrete stochastic latent $z_t$~\cite{hafner2021dreamerv2}.

\wmmodule{1}{Observation encoders}
At time $t$, $p_t$ is a short proprioceptive history of joint positions, joint velocities, base angular velocity, projected gravity, and previous actions. The depth image $d_t$ comes from a single head-mounted camera. Learned encoders $E_\phi$ map these inputs to the observation feature $e_t=E_\phi(p_t,d_t)$.

\wmmodule{2}{Recurrent transition}
Let $a_{t-1}$ represent the actions executed since the previous model update. The transition network $F_\phi$, implemented as a gated recurrent unit~\cite{cho2014gru}, advances the memory:
\begin{equation}
 h_t=F_\phi(h_{t-1},z_{t-1},a_{t-1}).
\label{eq:wm-transition}
\end{equation}
This update precedes incorporation of the current observation, so $h_t$ summarizes the preceding observation--action history.

\begin{subequations}\label{eq:wm-latent}
\wmmodule{3}{Prior head}
The neural network $p_\phi$ parameterizes a discrete prior distribution $P_t$ over the stochastic latent using $h_t$ alone. We denote a sample from this prior by $\hat z_t$:
\begin{equation}
 P_t=p_\phi(\,\cdot\mid h_t), \qquad \hat z_t\sim P_t.
\label{eq:wm-prior}
\end{equation}
\wmmodule{4}{Posterior head}
The neural network $q_\phi$ additionally uses $e_t$ to parameterize an observation-conditioned posterior $Q_t$ over the same latent space:
\begin{equation}
 Q_t=q_\phi(\,\cdot\mid h_t,e_t), \qquad z_t\sim Q_t.
\label{eq:wm-posterior}
\end{equation}
\end{subequations}
During observation-conditioned training and online interaction, the posterior sample $z_t$ is used in the next recurrent transition.

\wmblock{Observation and reward prediction}
Three prediction heads supervise the latent state $(h_t,z_t)$ so that it retains robot-state and visible-terrain information and encodes task-relevant outcomes. A proprioceptive decoder and a depth decoder reconstruct $p_t$ and $d_t$, respectively; a reward prediction head predicts the task reward. Together, these networks, denoted by $D_\phi$, map $[h_t,z_t]$ to reconstructions $\hat p_t$, $\hat d_t$ and reward estimate $\hat r_t$:
\begin{equation}
 (\hat p_t,\hat d_t,\hat r_t)=D_\phi(h_t,z_t).
\label{eq:wm-heads}
\end{equation}

\wmblock{World-model learning}
The world model learns from observation--action--reward sequences collected during policy training. Its training objective combines observation reconstruction, reward prediction, and prior--posterior matching to jointly learn latent states that preserve robot and terrain information and dynamics that predict action-conditioned evolution and task outcomes:
\begin{equation}
 \begin{aligned}
  \mathcal L_{\mathrm{WM},t}={}&\|s(\hat p_t)-s(p_t)\|_2^2\\
  &+\|\hat d_t-d_t\|_2^2+\mathcal L^r_t+\mathcal L^{\mathrm{KL}}_t.
 \end{aligned}
\label{eq:wm-loss}
\end{equation}
Here, $s(\cdot)$ denotes the elementwise symlog transform~\cite{hafner2023dreamerv3}. The squared errors train $D_\phi$ to reconstruct $p_t$ and $d_t$ from the posterior state $(h_t,z_t)$, preserving robot-state and visible-terrain information.

The loss $\mathcal L_t^r$ trains the reward prediction head to estimate the observed task reward $r_t$ as $\hat r_t$, encouraging the latent state to encode task-relevant outcomes~\cite{hafner2023dreamerv3}.

The Kullback--Leibler (KL) regularizer~\cite{hafner2023dreamerv3} matches the prior $P_t$ to the observation-conditioned posterior $Q_t$:
\begin{equation}
\begin{split}
\mathcal L_t^{\mathrm{KL}}={}&\beta_{\mathrm d}\max\!\bigl(D_{\mathrm{KL}}[\mathrm{sg}(Q_t)\|P_t],1\bigr)\\
&+\beta_{\mathrm r}\max\!\bigl(D_{\mathrm{KL}}[Q_t\|\mathrm{sg}(P_t)],1\bigr).
\end{split}
\label{eq:wm-kl}
\end{equation}
Here, $\mathrm{sg}$ stops gradients, and $\beta_{\mathrm d}=\klDyn{}$ and $\beta_{\mathrm r}=\klRep{}$ weight the two terms. The first term trains the action-conditioned prior against the posterior target; the second regularizes the posterior toward the prior. Since $P_t$ has no access to $e_t$, prior matching trains the dynamics to predict observation-informed states from preceding states and actions.

\wmblock{World model as a latent simulator}
These objectives train action-conditioned latent dynamics that can be propagated beyond the observed state. The multi-step predictions in \Cref{fig:wm-openloop} show that the world model can simulate future robot states and egocentric observations by recursively applying its learned dynamics to a supplied action sequence. The recurrent memory $h_t$ thus learns predictive structure about how actions affect future observations and rewards. Conditioning the actor on this memory gives it access to the learned relationship between motion and upcoming terrain, helping it generate actions that prepare the robot for the next foothold.

\subsection{Policy Conditioning and Joint Training}
At each control step, the WM-feature encoder $g_\psi$ (\Cref{fig:arch}), a projection network with parameters $\psi$, maps the most recently updated deterministic memory $h_t$ to the policy feature $f^{\mathrm{WM}}_t$. The actor--critic takes $(p_t,p_{t-H:t},c_t,f^{\mathrm{WM}}_t)$ as input, where $c_t$ is the commanded base-frame velocity. A history encoder embeds $p_{t-H:t}$, the proprioceptive inputs of the last $H=\histLen{}$ control steps. The actor $\pi_\theta$, with parameters $\theta$, outputs a joint-control action $a_t$:
\begin{equation}
 \begin{aligned}
  f^{\mathrm{WM}}_t&=g_\psi(h_t),\\
  a_t&\sim\pi_\theta(\,\cdot\mid p_t,p_{t-H:t},c_t,f^{\mathrm{WM}}_t).
 \end{aligned}
\label{eq:wm-policy}
\end{equation}
A mixture-of-experts (MoE) backbone fuses these inputs and feeds the actor and critic heads. Proprioception $p_t$ and command $c_t$ enter directly; depth enters only through $f^{\mathrm{WM}}_t$. The world model updates at 10~Hz and the policy at 50~Hz; $h_t$ and $f^{\mathrm{WM}}_t$ are held fixed between world-model updates.

The joint objective is $\mathcal L_{\mathrm{PPO}}+\mathcal L_{\mathrm{WM}}$, where $\mathcal L_{\mathrm{PPO}}$ is the PPO actor--critic loss and $\mathcal L_{\mathrm{WM}}$ averages \eqref{eq:wm-loss} over sampled model updates. Training starts jointly without world-model pretraining. World-model gradients also update the observation encoders. At deployment, the actor outputs actions directly from $(p_t,p_{t-H:t},c_t,f^{\mathrm{WM}}_t)$, without model rollout or action search.

\subsection{Training Rewards}
\label{sec:method:reward}

The standard locomotion reward $r_t^{\mathrm{std}}$ provides the basis for policy training. We augment it with contact-region rewards that discourage invalid foot placements and guide traversal of the available support regions. Let $\mathcal V_t^{\mathrm{foot}}$ denote the set of foot-mesh sample points transformed into the world frame~\cite{hikingwild2026}, and let $\mathbf 1[\cdot]$ denote an indicator.

\wmblock{Standard reward}
The standard reward $r_t^{\mathrm{std}}$ combines velocity and heading tracking with posture, gait, smoothness, energy, and joint-safety terms; \Cref{tab:reward-terms} summarizes the key terms and weights.

\wmblock{Stair-boundary penalty}
The stairs rise above the surrounding ground, which forms the penalty zones $\mathcal Z^{\mathrm{bdry}}$ outside the stair footprint. We count foot points projected into these zones:
\begin{equation}
r_t^{\mathrm{bdry}}=\sum_{\mathbf v\in\mathcal V_t^{\mathrm{foot}}}
\mathbf 1[\mathbf v_{xy}\in\mathcal Z^{\mathrm{bdry}}].
\label{eq:r-bdry}
\end{equation}
Here, $\mathbf v_{xy}$ is the horizontal projection of $\mathbf v$. A negative weight discourages detours around the stairs.

\wmblock{Riser-penetration penalty}
Let $\mathcal Z^{\mathrm{riser}}$ be the union of riser volumes, excluding a clearance near each tread edge. The violation count is
\begin{equation}
r_t^{\mathrm{riser}}=\sum_{\mathbf v\in\mathcal V_t^{\mathrm{foot}}}
\mathbf 1[\mathbf v\in\mathcal Z^{\mathrm{riser}}].
\label{eq:r-riser}
\end{equation}
A negative weight penalizes penetration while allowing valid oblique edge contacts.

\wmblock{Sequential stair-tread contact reward}
We reward contact with successive stair treads in order:
\begin{equation}
r_t^{\mathrm{tread}}=g_t\rho_t^{\mathrm{seq}},
\label{eq:r-tread}
\end{equation}
where $g_t$ indicates the current tread's activation window and $\rho_t^{\mathrm{seq}}$ is its alternation-weighted sole-contact ratio. Later treads activate only after sustained contact completes the current one.

\wmblock{Stepping-stone top-contact reward}
We reward contact with each stone's top surface:
\begin{equation}
r_t^{\mathrm{stone}}=\sum_{s\in\mathcal A_t}
g_{s,t}r_{s,t}^{\mathrm{top}},
\label{eq:r-stone}
\end{equation}
where $r_{s,t}^{\mathrm{top}}$ rewards centered top-surface contact and $g_{s,t}$ indicates the reward window. Each stone's window opens once per episode on sufficient contact and closes after a short duration or earlier if that contact is lost. The stone is then removed from the eligible set $\mathcal A_t$.
\par

\wmblock{Total reward}
The total reward $r_t$ at control step $t$ is
\begin{equation}
 \begin{aligned}
 r_t={}&r_t^{\mathrm{std}}+w_b r_t^{\mathrm{bdry}}+w_r r_t^{\mathrm{riser}}\\
       &+w_t r_t^{\mathrm{tread}}+w_s r_t^{\mathrm{stone}},
 \end{aligned}
 \label{eq:r-total}
\end{equation}
where \Cref{tab:reward-terms} lists the weights of these contact-region rewards for each training configuration; unused terms have zero weight.
\par

\section{Experiments}
\label{sec:Experimental}

We train WM-LOCO in Isaac Lab~\cite{isaaclab2025} with \numEnvs{} parallel Unitree G1 environments on \trainGpu{}. The simulated terrains include procedurally generated stairs, gaps, stepping stones, and balance beams. The policy uses proprioception and a single head-mounted depth stream, with control at 50~Hz and world-model updates at 10~Hz.

\subsection{Baseline Comparisons}
\label{sec:baseline-eval}

The PPO baseline shares our sensory inputs, rewards, proprioceptive encoder, adversarial motion prior~\cite{peng2021amp,escontrela2022amp}, MoE backbone, and training budget, but omits the world model and its auxiliary loss. We also compare with GLAD~\cite{fu2026glad} and CReF~\cite{hao2026cref}, which target similarly challenging terrain. GLAD uses global--local attention for terrain encoding of a robot-centric elevation map; CReF combines cross-modal attention with recurrent fusion of depth and proprioception.

We evaluate stairs, gaps, stepping stones, and balance beams at Easy, Medium, and Hard terrain difficulty levels. \Cref{app:terrains} specifies the terrain dimensions used in the evaluation. Terrain-level success rates average these three levels equally.

\begin{table}[!t]
\centering
\caption{Success rates (\%) for four methods and four terrains. Mean is the arithmetic average of Easy, Medium, and Hard. Bold denotes the best simulation result in each row.}
\label{tab:main}\label{tab:baselines}
\setlength{\tabcolsep}{1.5pt}
\renewcommand{\arraystretch}{1.03}
\footnotesize
\begin{tabular*}{\linewidth}{@{\extracolsep{\fill}}llccccc@{}}
\toprule
& & \multicolumn{4}{c}{Simulation} & Real\\
\cmidrule(lr){3-6}\cmidrule(l){7-7}
Terrain & Difficulty & PPO & GLAD & CReF & \shortstack{WM-\\LOCO} & \shortstack{WM-\\LOCO}\\
\midrule
\multirow{4}{*}{Stairs} & Easy & 87.0 & \textbf{95.3} & 95.0 & 94.2 & \multirow{4}{*}{90}\\
& Medium & 91.4 & 88.7 & 91.3 & \textbf{95.7} & \\
& Hard & 91.3 & 60.3 & 64.7 & \textbf{92.0} & \\
& Mean & 89.9 & 81.4 & 83.7 & \textbf{94.0} & \\
\midrule
\multirow{4}{*}{Gaps} & Easy & 0.0 & \textbf{98.0} & 93.7 & \textbf{98.0} & \multirow{4}{*}{90}\\
& Medium & 0.0 & 99.0 & 78.3 & \textbf{100.0} & \\
& Hard & 0.0 & 78.7 & 0.0 & \textbf{90.0} & \\
& Mean & 0.0 & 91.9 & 57.3 & \textbf{96.0} & \\
\midrule
\multirow{4}{*}{\shortstack[l]{Stepping\\stones}} & Easy & 0.0 & 76.0 & 94.3 & \textbf{97.1} & \multirow{4}{*}{100}\\
& Medium & 0.0 & 64.7 & 94.0 & \textbf{97.3} & \\
& Hard & 0.0 & 7.7 & 88.7 & \textbf{96.4} & \\
& Mean & 0.0 & 49.4 & 92.3 & \textbf{96.9} & \\
\midrule
\multirow{4}{*}{\shortstack[l]{Balance\\beam}} & Easy & 0.0 & \textbf{99.0} & 95.3 & \textbf{99.0} & \multirow{4}{*}{90}\\
& Medium & 0.0 & 90.7 & 94.7 & \textbf{96.7} & \\
& Hard & 0.0 & 0.0 & 41.1 & \textbf{80.7} & \\
& Mean & 0.0 & 63.2 & 77.0 & \textbf{92.1} & \\
\bottomrule
\end{tabular*}
\par\vspace{3pt}\raggedright
Real reports WM-LOCO hardware success at one fixed difficulty per terrain rather than separate Easy, Medium, and Hard tiers.
\par\vspace{6pt}
{\centering\figurecaptype
\includegraphics[width=\linewidth]{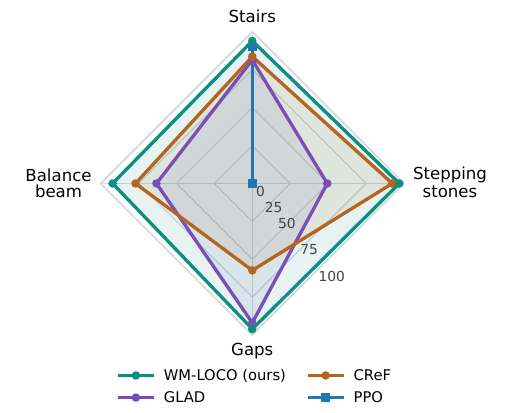}
\caption{Radar chart of success rates averaged across difficulty tiers. PPO achieves 0\% success on gaps, stepping stones, and balance beams.}
\label{fig:baselines}\par}
\end{table}

\subsection{World-Model Validation}

We test whether the world model propagates an observation-inferred state according to the supplied actions. \emph{Predict with actual actions} initializes the RSSM from depth and proprioception, then predicts 15 updates (1.5~s) ahead using the recorded actions without subsequent observations. Predicted depth, proprioception, and reward are compared with the corresponding Isaac Lab measurements. \emph{Posterior reconstruction} uses the current observation at each update to infer a posterior state and reconstruct that observation, without performing a multi-step rollout. The resulting reconstruction error serves as a decoding baseline for comparison with multi-step prediction errors.

Two further controls isolate temporal prediction and action dependence. \emph{Repeat initial observation} performs multi-step inference while reusing the initial observation as input at every update, providing a control for assessing whether the world model captures the temporal dynamics of the robot and its environment. \emph{Predict with other actions} substitutes a recorded action sequence from another time window, testing how the supplied actions affect the predicted state evolution.

\subsection{Ablation Study}

We conduct ablations on gap terrain to assess the contribution of each component. Three variants remove the world model, proprioceptive history encoder, or MoE, respectively. We vary the gap width to examine how each component affects traversal as the gap widens.

\begin{figure}[!t]
\centering
\includegraphics[width=\linewidth]{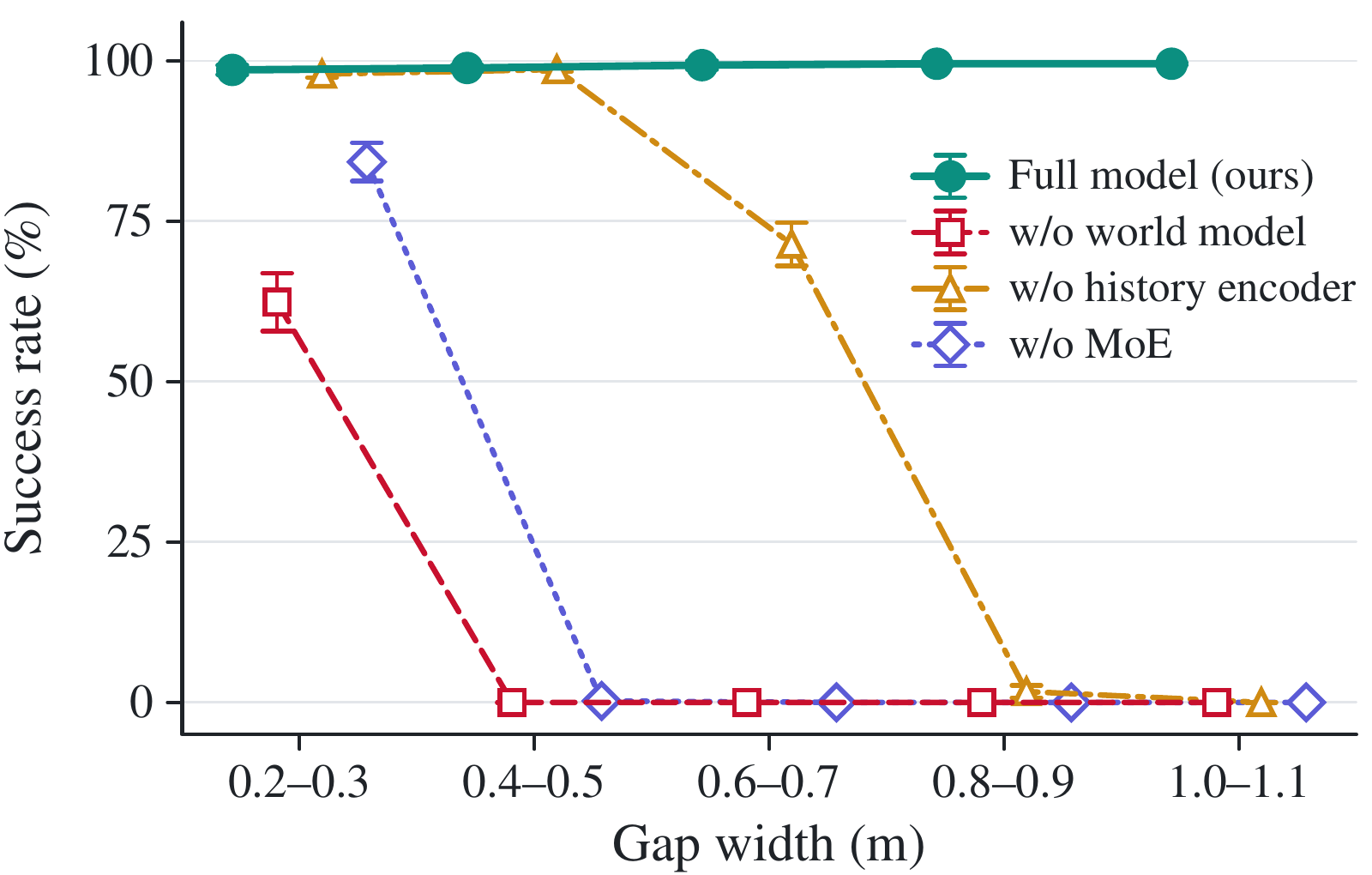}
\caption{Gap-traversal success of the full model and of variants without the world model, history encoder, or MoE.}
\label{fig:ablation}
\end{figure}

\subsection{Hardware Evaluation}
The policy runs onboard a Unitree G1 with a Jetson Orin. It uses proprioception and a single Intel RealSense D435 depth stream. We conduct ten trials each on stepping stones, stairs, a gap, and a balance beam. Stone tops are 0.25~m wide with 0.45~m gaps. Stairs have 0.15~m risers and 0.25~m treads. The gap spans 0.8~m. The balance beam is 3~m long and 0.10~m wide.

\section{Results}
\label{sec:results}

\subsection{Baseline Comparisons}
\paragraph{Success rates}
Averaged across the three terrain difficulty levels, WM-LOCO achieves 96.9\% success on stepping stones, 96.0\% on gaps, and 92.1\% on balance beams, compared with 0\% for the PPO baseline on all three terrains (\Cref{tab:main}). On stairs, success increases from 89.9\% to 94.0\%.
World-model augmentation thus enables reliable traversal where the PPO baseline records no successful trials, while retaining high stair success. On a combined course that chains all four terrain classes in one episode (\Cref{fig:realworld_teaser}(d)), WM-LOCO achieves 95.4\% success.

\paragraph{Gait quality}
On stairs, both methods achieve high success, but WM-LOCO produces more efficient and smoother motion. Stride length increases by 25.6\%, and pelvis acceleration decreases by 28.9\%. These percentages are mean relative changes across the three tiers. Steps per meter and normalized mechanical energy also decrease at every tier (\Cref{tab:stairs-gait}).

\paragraph{Comparison with GLAD and CReF}

WM-LOCO is the only method whose mean success exceeds 90\% on every terrain class (\Cref{fig:baselines}). Its four-terrain mean is 94.8\%, compared with 71.5\% for GLAD and 77.6\% for CReF. The baselines exhibit distinct weaknesses: GLAD degrades on stepping stones and beams, whereas CReF degrades most strongly on gaps.

Finite elevation-map sampling and feature downsampling in GLAD may reduce the precision of narrow support boundaries~\cite{fu2026glad}. CReF's recurrent fusion of depth and proprioception~\cite{hao2026cref} may likewise leave the takeoff-and-landing dynamics of wide gaps insufficiently constrained.

\subsection{World-Model Validation}

The model predicts evolving depth more accurately than either repeating the initial observation or supplying alternative actions throughout the evaluated 1.5-s horizon (\Cref{fig:wm-openloop}). This advantage supports the model's ability to predict temporal dynamics from the supplied actions. At 1.5~s, joint-position and joint-velocity errors remain close to the current-observation reconstruction reference. Reward prediction reaches a mean absolute error (MAE) of 0.072 and a correlation of 0.887 at 1.0~s.

Together, these results validate multi-step simulation of robot state, egocentric geometry, and task outcomes on the evaluated stepping-stone trajectories. The RSSM therefore learns predictive structure beyond current-observation reconstruction. Its recurrent feature provides the actor with predictive context about body motion and upcoming support, which can inform anticipatory foot placement.

\Needspace{3\baselineskip}
\subsection{Ablation Results}

The complete model achieves 98.6--99.6\% success across the tested gap widths (\Cref{fig:ablation}). Removing the world model reduces success to 0\% for every tested width range from 0.4--0.5~m onward. This collapse identifies world-model augmentation as essential to reliable gap traversal in the proposed system and supports the effectiveness of supplying its learned predictive context to the policy. The other ablations show how policy components help use this context: removing MoE also produces near-total failure beyond the smallest gap range, whereas removing the history encoder mainly impairs wider-gap traversal.

\subsection{Sim-to-Real Transfer}
Without additional tuning, WM-LOCO succeeds in 10/10 stone trials, 9/10 stair trials, 9/10 gap trials, and 9/10 beam trials. These results demonstrate sim-to-real transfer using only onboard depth and proprioception, with 37/40 successful trials overall (92.5\%).

\begin{table}[!t]
\centering
\caption{Terrain dimensions in simulation and on the real course.}
\label{tab:tier-dims}
\footnotesize
\setlength{\tabcolsep}{1.5pt}
\renewcommand{\arraystretch}{1.08}
\begin{tabular*}{\linewidth}{@{\extracolsep{\fill}}lcccc@{}}
\toprule
Dimension & \diffEasy{} & \diffMed{} & \diffHard{} & Real \\
\midrule
\multicolumn{5}{@{}l}{\emph{Stairs}} \\
Riser height & \stairRiserE{}\,cm & \stairRiserM{}\,cm & \stairRiserH{}\,cm & \realStairRiserCm{}\,cm \\
Tread depth & \stairTreadE{}\,cm & \stairTreadM{}\,cm & \stairTreadH{}\,cm & \realStairTreadCm{}\,cm \\
Number of steps & \stairStepsE{} & \stairStepsM{} & \stairStepsH{} & \realStairStepsPerFlight{} \\
\midrule
\multicolumn{5}{@{}l}{\emph{Gap}} \\
Gap width & \gapWidthEasy{}\,m & \gapWidthMed{}\,m & \gapWidthHard{}\,m & 0.80\,m \\
\midrule
\multicolumn{5}{@{}l}{\emph{Stepping stones}} \\
Stone edge & \stoneEdgeE{}\,cm & \stoneEdgeM{}\,cm & \stoneEdgeH{}\,cm & \realStoneEdgeCm{}\,cm \\
Same-lane gap & \stoneGapE{}\,cm & \stoneGapM{}\,cm & \stoneGapH{}\,cm & \realStoneGapCm{}\,cm \\
\midrule
\multicolumn{5}{@{}l}{\emph{Balance beam}} \\
Beam width & $29.6\text{--}32.3$\,cm & $18.8\text{--}21.5$\,cm & $8.0\text{--}10.7$\,cm & 10.0\,cm \\
Beam length & $2.15\text{--}2.30$\,m & $2.75\text{--}2.90$\,m & $3.35\text{--}3.50$\,m & 3.00\,m \\
\bottomrule
\end{tabular*}
\end{table}

\begin{table}[!t]
\centering
\caption{Key reward terms and weights. ``---'' = not used or inactive on that terrain.}
\label{tab:reward-terms}
\footnotesize
\setlength{\tabcolsep}{3pt}
\renewcommand{\arraystretch}{1.0}
\begin{tabular*}{\linewidth}{@{\extracolsep{\fill}}lcc@{}}
\toprule
Term & Stairs & Stones \\
\midrule
\multicolumn{3}{@{}l}{\emph{Standard locomotion ($r_t^{\mathrm{std}}$), selected terms}} \\
\multicolumn{3}{@{}l}{\emph{Tracking and progress}} \\
Linear-velocity tracking & $+3.0$ & $+3.0$ \\
Angular-velocity tracking & $+3.0$ & $+3.0$ \\
Heading-error penalty & $-1.0$ & $-1.0$ \\
No-progress penalty & $-2.0$ & $-2.0$ \\
Alive bonus & $+0.5$ & $+0.5$ \\
\midrule
\multicolumn{3}{@{}l}{\emph{Posture and gait}} \\
Pose-anchored stand-still (exp.) & $+1.5$ & --- \\
Air-time bonus & $+0.5$ & $+0.5$ \\
Foot-slide penalty & $-0.4$ & $-0.4$ \\
Foot flat-orientation penalty & $-0.4$ & $-0.4$ \\
Flat-orientation $L_2$ & $-3.0$ & $-3.0$ \\
\midrule
\multicolumn{3}{@{}l}{\emph{Smoothness and energy}} \\
Walking-gated action-rate $L_2$ & $-0.005$ & $-0.005$ \\
Joint torque $L_2$ & $-1.5{\times}10^{-7}$ & $-1.5{\times}10^{-7}$ \\
Joint acceleration $L_2$ & $-1.25{\times}10^{-7}$ & $-1.25{\times}10^{-7}$ \\
Mechanical-energy penalty & $-5{\times}10^{-5}$ & $-5{\times}10^{-5}$ \\
\midrule
\multicolumn{3}{@{}l}{\emph{Joint and contact safety}} \\
Joint-position soft limit & $-1.0$ & $-1.0$ \\
Joint-velocity soft limit & $-1.0$ & $-1.0$ \\
Torque soft limit & $-0.01$ & $-0.01$ \\
Undesired-contact penalty & $-1.0$ & $-1.0$ \\
\midrule
\multicolumn{3}{@{}l}{\emph{Contact-region rewards in \eqref{eq:r-total}}} \\
Stair boundary ($r_t^{\mathrm{bdry}}$) & $-0.05$ & --- \\
Riser penetration ($r_t^{\mathrm{riser}}$) & $-0.5$ & --- \\
Sequential tread contact ($r_t^{\mathrm{tread}}$) & $+0.5$ & --- \\
Stone top contact ($r_t^{\mathrm{stone}}$) & --- & $+4.0$ \\
\bottomrule
\end{tabular*}
\end{table}

\section{Limitations}
\label{sec:limitations}

\textbf{Scope.} Evaluation covers discrete footholds and narrow support regions. Generalization to stepping stones with varying elevations and outdoor unstructured terrain remains untested.
\textbf{Platform.} Experiments use only Unitree G1; transfer to other humanoid robot platforms has not been evaluated.
\textbf{Visual perception.} The policy relies on depth input. Future work will explore stereo RGB input toward broader locomotion and manipulation capabilities.

\section{Conclusion}
\label{sec:conclusion}

WM-LOCO couples a world model with PPO to learn predictive context for visual humanoid locomotion. Multi-step validation demonstrates that the model simulates robot-state and egocentric scene evolution without future observations. Ablation results support the effectiveness of the world-model augmentation and complementary policy components. WM-LOCO reliably traverses gaps, stepping stones, and balance beams where the PPO baseline fails, while improving stair gait efficiency and smoothness. Zero-shot sim-to-real transfer achieves 92.5\% success across stepping stones, stairs, gaps, and balance beams. Together, these results demonstrate the effectiveness of world-model-augmented control for reliable humanoid locomotion on foothold-constrained terrain.

\appendices
\renewcommand{\thesection}{\Roman{section}}

\crefalias{section}{appendix}

\begin{figure}[!t]
\centering
\includegraphics[width=\linewidth]{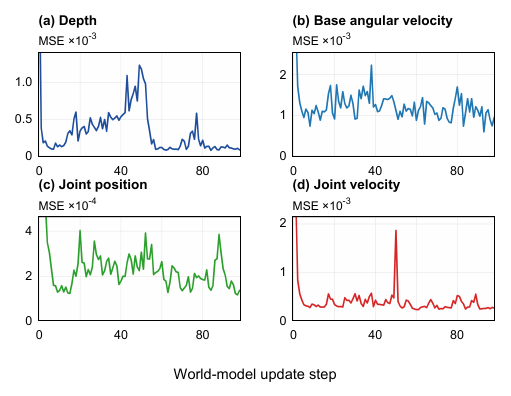}
\caption{Posterior reconstruction errors for depth and robot-state measurements.}
\label{fig:wm-recon}
\end{figure}

\begin{table}[!t]
\centering
\caption{Stair gait, energy, and smoothness metrics per
difficulty tier. Arrows indicate the preferred direction; bold denotes
the better value in each row that has a preferred direction.}
\label{tab:stairs-gait}
\footnotesize
\setlength{\tabcolsep}{2.5pt}
\begin{tabular*}{\linewidth}{@{\extracolsep{\fill}}p{0.32\linewidth}llcc@{}}
\toprule
Metric & Unit & Difficulty & \ppo & \wmppo \\
\midrule
\multirow{3}{=}{Stride length$\uparrow$} & \multirow{3}{*}{m}
   & \diffEasy{} & \stairStrideEasyPPO & \textbf{\stairStrideEasyWM} \\
&  & \diffMed{}  & \stairStrideMedPPO  & \textbf{\stairStrideMedWM}  \\
&  & \diffHard{} & \stairStrideHardPPO & \textbf{\stairStrideHardWM} \\
\midrule
\multirow{3}{=}{Steps per meter$\downarrow$} & \multirow{3}{*}{$\mathrm{m}^{-1}$}
   & \diffEasy{} & \stairSpmEasyPPO & \textbf{\stairSpmEasyWM} \\
&  & \diffMed{}  & \stairSpmMedPPO  & \textbf{\stairSpmMedWM}  \\
&  & \diffHard{} & \stairSpmHardPPO & \textbf{\stairSpmHardWM} \\
\midrule
\multirow{3}{=}{Normalized mechanical energy$\downarrow$} & \multirow{3}{*}{---}
   & \diffEasy{} & \stairCotEasyPPO & \textbf{\stairCotEasyWM} \\
&  & \diffMed{}  & \stairCotMedPPO  & \textbf{\stairCotMedWM}  \\
&  & \diffHard{} & \stairCotHardPPO & \textbf{\stairCotHardWM} \\
\midrule
\multirow{3}{=}{Pelvis acc.$\downarrow$} & \multirow{3}{*}{m/s$^2$}
   & \diffEasy{} & \stairPelaccEasyPPO & \textbf{\stairPelaccEasyWM} \\
&  & \diffMed{}  & \stairPelaccMedPPO  & \textbf{\stairPelaccMedWM}  \\
&  & \diffHard{} & \stairPelaccHardPPO & \textbf{\stairPelaccHardWM} \\
\midrule
\multirow{3}{=}{Action-rate$\downarrow$} & \multirow{3}{*}{---}
   & \diffEasy{} & \stairAraEasyPPO & \textbf{\stairAraEasyWM} \\
&  & \diffMed{}  & \stairAraMedPPO  & \textbf{\stairAraMedWM}  \\
&  & \diffHard{} & \stairAraHardPPO & \textbf{\stairAraHardWM} \\
\midrule
\multirow{3}{=}{Torque/limit} & \multirow{3}{*}{---}
   & \diffEasy{} & \stairTqEasyPPO & \stairTqEasyWM \\
&  & \diffMed{}  & \stairTqMedPPO  & \stairTqMedWM  \\
&  & \diffHard{} & \stairTqHardPPO & \stairTqHardWM \\
\bottomrule
\end{tabular*}
\end{table}

\section{Terrain Generation Parameters}
\label{app:terrains}
\Cref{tab:tier-dims} lists the physical terrain dimensions in simulation and on the real course. Stepping stones have a fixed same-lane pitch of $0.70\,\mathrm{m}$. The real balance beam is \realBeamLength{} long and \realBeamWidth{} wide.

\section{Reward Terms}
\label{app:reward}
\Cref{tab:reward-terms} summarizes the key reward terms and weights
for the gap-and-stair (\emph{Stairs}) and stepping-stone (\emph{Stones})
training configurations. The standard terms contribute to $r_t^{\mathrm{std}}$;
the four contact-region rewards correspond to those described in
\Cref{sec:method:reward} and combined in \eqref{eq:r-total}.
Auxiliary shaping and regularization terms are omitted for brevity.

\Needspace{11\baselineskip}
\section{Posterior Reconstruction}
\label{app:wm-openloop}

\Cref{fig:wm-recon} reports posterior reconstruction mean squared error (MSE) for depth and robot-state measurements. At each update, the model encodes and decodes the current observation without a multi-step rollout. Curves show median MSE across seven parallel environments, with depth normalized to $[0,1]$. Errors remain low overall, with transient increases in depth and joint velocity, indicating that the latent state preserves information about the current observation.

\section{Stair Gait Metrics by Difficulty}
\label{app:stairs-gait}
\Cref{tab:stairs-gait} reports per-difficulty gait, energy,
and smoothness metrics on \emph{stairs}, where the two methods
achieve similar success rates. Longer strides and fewer steps per meter indicate a more efficient
gait. Lower normalized mechanical energy indicates improved energy efficiency, while lower pelvis acceleration and action rate indicate smoother motion. Torque relative to the actuator limit is also reported.

\end{document}